\documentclass[11pt]{article}

\PassOptionsToPackage{table}{xcolor}
\usepackage{acl}

\usepackage{times}
\usepackage{latexsym}
\usepackage[T1]{fontenc}
\usepackage[utf8]{inputenc}
\usepackage{microtype}
\usepackage{url}
\usepackage{xurl}
\usepackage{booktabs}
\usepackage{amsfonts}
\usepackage{amssymb}
\usepackage{nicefrac}
\usepackage{graphicx}
\usepackage{tikz}
\usepackage{float}
\usepackage{wrapfig}
\usepackage{array}
\usepackage{adjustbox}
\usepackage{caption}
\usepackage{cuted}
\usepackage{mmstyles}
\usepackage{cleveref}

\newcommand{\cmark}{\ensuremath{\checkmark}}

\newcommand{\xmark}{\ensuremath{\times}}
\definecolor{tablegroupblue}{HTML}{F2F8FC}
\definecolor{vmlnavy}{HTML}{06184F}
\definecolor{vmllblue}{HTML}{1687FF}
\definecolor{vmlcyan}{HTML}{00B7D7}
\definecolor{vmlgreen}{HTML}{18B86A}
\definecolor{vmlolive}{HTML}{AFCB16}
\definecolor{vmlorange}{HTML}{FF9D00}
\definecolor{vmlcoral}{HTML}{FF6846}
\definecolor{vmlpink}{HTML}{F13D66}

\newcommand{\vmltitleword}{%
  \texorpdfstring{%
    \textbf{\textit{%
      \textcolor{vmlnavy}{Video-MME-}%
      \textcolor{vmllblue}{L}%
      \textcolor{vmlcyan}{o}%
      \textcolor{vmlgreen}{g}%
      \textcolor{vmlolive}{i}%
      \textcolor{vmlorange}{c}%
      \textcolor{vmlcoral}{a}%
      \textcolor{vmlpink}{l}%
    }}%
  }{Video-MME-Logical}%
}

\newcommand{\vmltitleicon}{%
  \texorpdfstring{%
    \raisebox{-0.32em}{\includegraphics[height=1.55em]{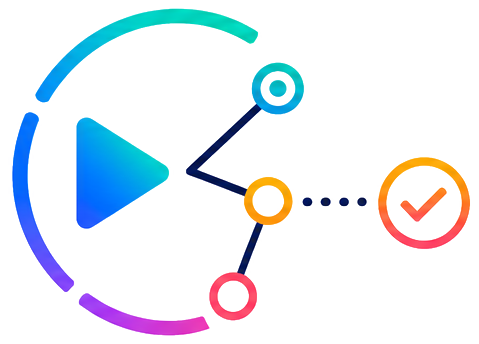}}\hspace{0.25em}%
  }{}%
}

\title{\vmltitleicon\vmltitleword: A Controlled Diagnostic Benchmark for \\ Video Temporal-Logical Reasoning}

\author{%
  \normalfont Hohin Kwan\textsuperscript{1*}, Hongyu Li\textsuperscript{2*\ensuremath{\dagger}}, Ray Zhang\textsuperscript{3}, Manyuan Zhang,\\
  \normalfont Xianghao Kong\textsuperscript{1}, Anyi Rao\textsuperscript{1}, Jiahao Xie\textsuperscript{2\ensuremath{\ddagger}}, Si Liu\textsuperscript{2}\\
  \normalfont\small \textsuperscript{1}HKUST \quad \textsuperscript{2}Colab, Beihang University \quad \textsuperscript{3}CUHK\\
  \normalfont\small Project page: \url{https://mrakas.github.io/video-mme-logical/}\\
  \normalfont\small \textsuperscript{*}Equal contribution. \textsuperscript{\ensuremath{\dagger}}Project Leader. \textsuperscript{\ensuremath{\ddagger}}Corresponding Author.
}

\begin{document}

\maketitle

\begin{strip}
\vspace{-3.8em}
\centering
\includegraphics[width=\textwidth]{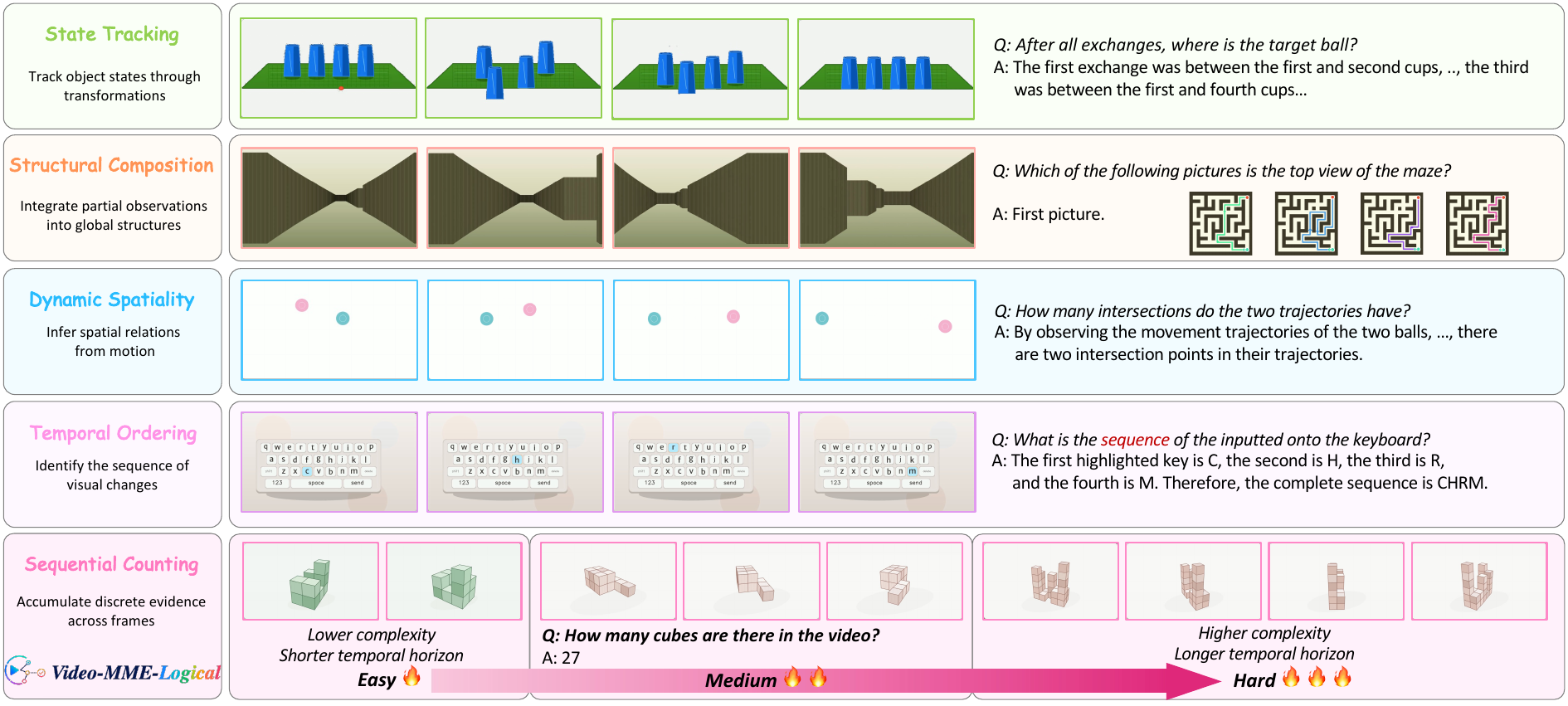}
\captionof{figure}{
\textsc{Video-MME-Logical} is a controllable benchmark for video temporal-logical reasoning with 25 tasks. It evaluates whether models can reason over dynamic visual worlds through \textcolor[HTML]{7CCB4D}{State Tracking}, \textcolor[HTML]{FF8B4A}{Structural Composition}, \textcolor[HTML]{36BDF2}{Dynamic Spatiality}, \textcolor[HTML]{EA7AD8}{Temporal Ordering} and \textcolor[HTML]{FF6FB5}{Sequential Counting}, spanning final-answer tasks, intermediate-state diagnostics, and difficulty-controlled settings.}
\label{fig:teaser}
\end{strip}

\begin{abstract}

Recent interest in multimodal large language models (MLLMs) raises a central question: can they reason over dynamic visual evidence rather than merely recognize objects or events in individual frames? This ability, which we refer to as \emph{video temporal-logical reasoning}, requires models to maintain, update, and compose evidence as visual states evolve across frames. Existing video benchmarks often conflate this capability with scene complexity, static recognition, or uncontrolled temporal variation. To isolate this capability, we introduce \textsc{Video-MME-Logical}, a controlled benchmark organized around five temporal-logical operations: state tracking, sequential counting, temporal ordering, dynamic spatiality, and structural composition. The benchmark contains 25 fine-grained task categories generated with controlled object states, transitions, temporal dependencies, and logical compositions. It enables difficulty-controlled final-answer evaluation by varying temporal horizon and reasoning complexity, and supports intermediate-state diagnostics by verifying whether models recover the required logical reasoning trace before producing the final answer. Experiments with state-of-the-art MLLMs reveal a substantial human-model gap, especially as temporal-logical complexity increases. Supervised fine-tuning on up to 500K generated samples improves performance but remains insufficient to close the reasoning gap, positioning \textsc{Video-MME-Logical} as a scalable testbed for analyzing and improving temporal-logical reasoning in MLLMs.
\end{abstract}

\section{Introduction}

Human visual cognition inherently relies on temporal-logical reasoning---the capacity to synthesize shifting visual sequences by maintaining, updating, and composing evidence as an event unfolds over time~\cite{kahneman1992reviewing,pylyshyn2001visual,cavanagh2011visual}.
Currently, the quest to replicate this human ability is dominated by multimodal large language models (MLLMs)~\cite{google2025gemini3pro, openai2026gpt54, liu2024deepseek, team2026qwen3, glm, li2025reinforcement}, which have achieved impressive performance across standard video understanding benchmarks~\citep{videomme, videoreasonbench, videomme2026, tempcompass, vstar}.
Despite their remarkable advancements, a critical conceptual gap remains: merely aggregating multi-frame inputs is not equivalent to executing logical reasoning over time. For instance, while a young child can effortlessly deduce the position of a hidden ball in a shell game by tracking motion through occlusion, contemporary MLLMs struggle with this foundational logic.
Since existing benchmarks often conflate basic frame-stitching with robust temporal inference, the true logical faculties of current video models remain heavily overestimated.

This gap persists primarily because existing benchmarks fail to provide a controlled diagnostic setting that isolates temporal-logical reasoning from general temporal understanding.
We identify three key limitations in existing evaluations: (1) The \textbf{categories} of temporal-logical reasoning remain under-specified. Many existing benchmarks are typically organized by data source, scene type, event category, or action class, rather than by temporal-logic operations, making it difficult to attribute model errors to specific reasoning capabilities. (2) Existing temporal benchmarks struggle to provide interpretable \textbf{difficulty levels}, since difficulty in natural videos often co-varies with scene complexity, annotation language, and dataset bias rather than  controlled by temporal dependencies and logical complexity. (3) Most existing benchmarks evaluate only final answers and lack \textbf{verifiable intermediate states}, making it difficult to determine whether a model truly performs temporal-logical reasoning or merely relies on local cues.

To address these limitations, we propose \textsc{Video-MME-Logical}, a controlled diagnostic benchmark for video temporal-logical reasoning. Our design targets the three gaps above. \textbf{First}, to make temporal-logical reasoning explicit rather than loosely defined, we organize the benchmark around five temporal-logical operations: State Tracking, Sequential Counting, Temporal Ordering, Dynamic Spatiality, and Structural Composition. These operations specify what information must be maintained, accumulated, ordered, spatially inferred, or composed over time, turning a broad notion of video reasoning into an operation-centric evaluation framework. \textbf{Second}, to make difficulty interpretable and controllable, we construct 25 fine-grained task categories through procedural generation, which allows us to precisely control object states, state transitions, temporal dependencies, and logical compositions. Each task category is further divided into three difficulty levels according to temporal horizon and reasoning complexity. \textbf{Third}, to move beyond final-answer-only evaluation, we introduce \textsc{Video-MME-Logical-S}, an intermediate-state diagnostic subset whose intermediate evidence can be described and verified.

To study the effect of data scaling on video temporal-logical reasoning, we generate 500K procedurally created training samples and fine-tune Qwen3-VL-8B~\citep{qwen3vl} with different data scales. The results show that supervised fine-tuning (SFT) improves performance, reaching 40\% accuracy at 375K samples, but further scaling does not provide clear additional gains. A substantial gap to human experts remains, suggesting that data scaling alone is insufficient and that current models may still struggle with long temporal horizons and complex logical structures. We hope \textsc{Video-MME-Logical}, with its scalable training data and diagnostic evaluation setting, can support future research on video temporal-logical reasoning.

In summary, our contributions are as follows:
\begin{itemize}
 \item We propose a taxonomy of temporal-logical reasoning and instantiate it as \textsc{Video-MME-Logical}, a controlled benchmark with five operation categories and 25 fine-grained tasks.
\item We design difficulty-controlled evaluation settings and intermediate-state diagnostics, revealing a substantial human-model gap in temporal-logical reasoning.
\item We build a large-scale training set and conduct scaling studies, showing that more data improves performance but remains insufficient to close the reasoning gap.
\end{itemize}

\section{Related Work}

    \begin{table*}[h]
  \centering
  \footnotesize
  \setlength{\tabcolsep}{4pt}
  \begin{tabular}{@{}lrrrrccc@{}}
    \toprule
    Benchmark & \#Tasks & \#Videos & \#Train & \#Test & Control. & Difficulty. & Intermediate. \\
    \midrule
    TOMATO~\citep{tomato} & 6 & 1,417 & 0 & 1,417 & \xmark & \xmark & \xmark \\
    TempCompass~\citep{tempcompass} & 5 & 410 & 0 & 410 & \xmark & \xmark & \xmark \\
    ReXTime~\citep{rextime} & 3 & 12,759 & 9,695 & 3,064 & \xmark & \xmark & \xmark \\
    V-STaR~\citep{vstar} & 2 & 2,094 & 0 & 2,094 & \xmark & \xmark & \xmark \\
    \textsc{Video-MME-Logical} & 25 & 503,750 & 500,000 & 3,750 & \cmark & \cmark & \cmark \\
    \bottomrule
  \end{tabular}
  \caption{Comparison with recent video temporal reasoning benchmarks. Control. indicates whether the benchmark can programmatically control each element in the video; Difficulty. indicates whether it provides controlled difficulty settings; Intermediate. indicates whether it supports verifiable intermediate-state evaluation.}
  \label{tab:benchmark-comparison}
\end{table*}

\noindent\textbf{General Video Understanding Benchmarks.}
Video-language benchmarks have evolved from short-clip question answering and action-level understanding toward broad evaluations of general video comprehension. Recent suites such as Video-MME~\citep{videomme}, MLVU~\citep{mlvu}, ALLVB~\citep{allvb}, LongVideoBench~\citep{longvideobench}, LVBench~\citep{lvbench}, CinePile~\citep{cinepile}, VideoEspresso~\citep{han2025videoespresso} and MovieChat~\citep{moviechat} expand evaluation along multiple general-purpose axes, including video domains, durations, task formats, and open-ended instruction following. Other benchmarks improve realism or diagnostic coverage by focusing on egocentric activities, temporal relations, object interactions, and perception-oriented video understanding, as in Ego4D~\citep{ego4d}, EgoSchema~\citep{egoschema}, MVBench~\citep{mvbench}, NExT-QA~\citep{nextqa}, and the Perception Test~\citep{perceptiontest}. Together, these benchmarks are valuable precisely because they approximate diverse natural video-use scenarios and measure broad multimodal competence.

This breadth, however, leaves a more specific question under-specified: whether models can perform well-defined temporal-logical operations under controlled visual conditions. General video benchmarks are intentionally heterogeneous, and thus rarely organize questions around a fixed set of logical categories or difficulty-controlled traces. In contrast, \textsc{Video-MME-Logical} uses synthetic, controllable videos to isolate temporal-logical reasoning across multiple dimensions.

\noindent\textbf{Video Temporal Reasoning Benchmarks.}
Recent benchmarks have moved beyond broad video comprehension toward targeted evaluations of temporal and spatiotemporal reasoning. VITATECS~\citep{vitatecs} probes fine-grained temporal concepts through counterfactual caption discrimination; TemporalVQA~\citep{temporalvqa} studies temporal order and time-lapse reasoning from image pairs; and TempCompass~\citep{tempcompass} covers action, speed, direction, attribute change, and event order. Other benchmarks emphasize temporally grounded evidence: TOMATO~\citep{tomato} evaluates whether models can use ordered observations from continuous frames, ReXTime~\citep{rextime} links questions and answers across video segments, V-STaR~\citep{vstar} studies chains that connect what, when, and where information, and VideoReasonBench~\citep{videoreasonbench} evaluates vision-centric complex video reasoning. Together, these benchmarks show that current models still struggle with temporal cues, event ordering, cross-time relations, and spatiotemporal grounding.

Recent efforts have begun to examine video temporal-logical reasoning. \citet{cups} focuses on shell-game-style tracking, offering a controlled but task-specific test of logical reasoning, whereas \citet{verybig} studies video-generation-oriented reasoning at scale and emphasizes broad video-reasoning behaviors. These studies reveal important failures in temporal tracking and video reasoning, but they do not systematically distinguish which temporal-logical operation an MLLM must execute to solve a problem, such as maintaining hidden states, accumulating evidence, ordering visual changes, inferring dynamic spatial relations, or composing partial observations. As a result, final-answer accuracy alone makes it difficult to determine whether a model truly performs temporal-logical reasoning. \textsc{Video-MME-Logical} addresses this gap by providing a comprehensive MLLM benchmark organized around five temporal-logical operations, with controllable difficulty and verifiable intermediate states; \Cref{tab:benchmark-comparison} summarizes these distinctions.

\section{\textsc{Video-MME-Logical} Benchmark}

This section presents the construction of \textsc{Video-MME-Logical}, including its task architecture, dataset statistics, and data curation pipeline.

\subsection{Temporal-Logical Reasoning Taxonomy}

\begin{figure*}[t]
  \centering
  \includegraphics[width=\textwidth]{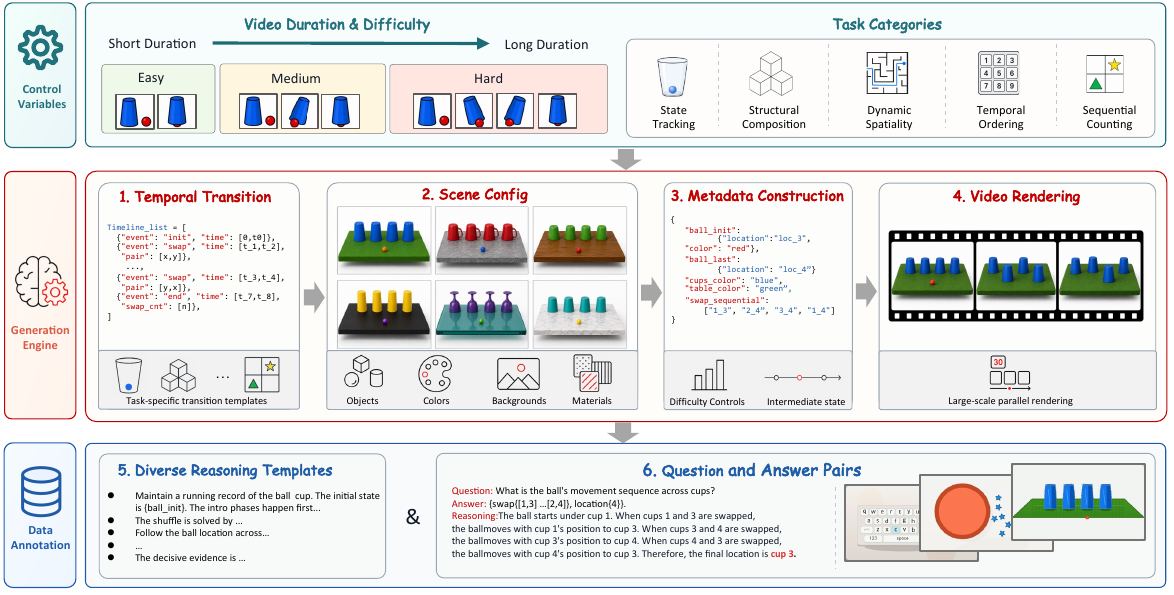}
  \caption{\textsc{Video-MME-Logical} combines controllable video generation, structured metadata, and diversified reasoning templates to build a 25-task temporal logical reasoning benchmark.}
  \label{fig:pipeline}

\end{figure*}

We organize \textsc{Video-MME-Logical} around five foundational temporal-logical reasoning abilities. These categories are motivated by the observation that solving a video reasoning problem requires more than recognizing visible objects or events: a model must decide what information should be remembered, how it should be updated, and how multiple temporal states should be composed to support inference. Specifically, \textbf{State Tracking} refers to maintaining latent or hidden object states across visual transformations, especially when the target state is no longer directly visible. \textbf{Sequential Counting} refers to accumulating discrete evidence over time, where the answer depends on a temporal history rather than any individual frame. \textbf{Temporal Ordering} refers to identifying the order of state changes, revealed symbols, or event sequences that determine the final outcome. \textbf{Dynamic Spatiality} refers to inferring geometric and dynamic relations from continuous movement, including trajectories, rotations, intersections, and relative speeds. \textbf{Structural Composition} refers to composing spatial structures across viewpoints, occlusions, and partial observations. To operationalize these abilities, each category is implemented as a set of parameterized task generators that produce videos, questions, answers, and, when applicable, verifiable intermediate reasoning traces. Representative examples of these abilities are shown in Fig.~\ref{fig:pipeline}.

\subsection{Data Statistics}
\textsc{Video-MME-Logical} contains 503,750 videos in total, including 500K training videos and 3,750 test videos. The test set covers 25 fine-grained task categories organized into five temporal-logical operation groups. Each category is evaluated in three difficulty levels (\ie, \textit{easy}, \textit{medium}, and \textit{hard}), which are defined by category-specific temporal horizons and reasoning complexity factors. Within the 25-category task space, an 8-category intermediate-state diagnostic subset provides intermediate-state annotations to verify whether the models maintain the correct temporal evidence trace rather than only producing the final answer. Detailed task definitions and examples are provided in Appendix~\ref{app:task-details}.

\begin{figure}[t]
  \centering
  \includegraphics[width=\linewidth]{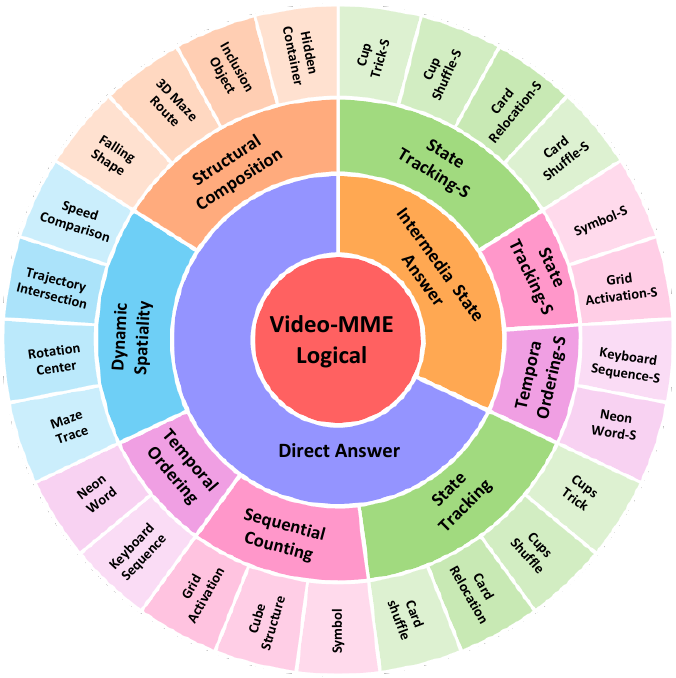}

  \caption{Taxonomy of \textsc{Video-MME-Logical}. The inner ring separates direct-answer tasks from the intermediate-state diagnostic subset, while the outer rings group fine-grained task categories under five temporal-logical operation groups.}
  \vspace{-4mm}
  \label{fig:taxonomy-pie}
\end{figure}

\subsection{Programmatic Generation Pipeline}
\label{sec:programmatic-generation}
Inspired by synthetic benchmarks such as CLEVR~\citep{johnson2017clevr,  yi2020clevrer, zhuo2025factuality}, \textsc{Video-MME-Logical} uses programmatic generation to decouple temporal-logical reasoning from visual noise and annotation ambiguity in natural videos. This design enables reproducible task generation, controllable difficulty, and verifiable intermediate evidence.

\noindent
\textbf{Task Program Design.}
Each task category is implemented as an executable program composed of four core components: 
(1) \textit{temporal transition}, which defines how the scene evolves over time, including the order in which objects move, swap, appear, or disappear; 
(2) \textit{scene configuration}, which specifies the objects, visual attributes, spatial layout, and optional distractors in the video; 
(3) \textit{metadata construction}, which records the complete temporal evidence trace for video rendering, question generation, answer computation, difficulty control, and intermediate-state supervision;
and (4) \textit{video rendering}, which converts the recorded metadata into an MP4 video at 30 FPS.

\noindent
\textbf{Difficulty Control.}
\textsc{Video-MME-Logical} defines three difficulty levels, \ie, \textit{easy}, \textit{medium}, and \textit{hard}, by increasing both temporal horizon and reasoning complexity. In cup-tracking tasks, for example, video duration instantiates temporal horizon, increasing from 10 seconds in \textit{easy} videos to 20 seconds in \textit{hard} videos, while the number of swaps instantiates reasoning complexity, increasing from 4 to 8 swaps.

\noindent
\textbf{Intermediate-State Annotation.}
For task categories with intermediate-state diagnostics, the program metadata provides supervision beyond the final answer. For example, in cup-tracking tasks, the metadata records the full state trajectory, including the initial ball position, each cup-swap pair with its timestamp, and the final ball position. These annotations allow us to evaluate whether a model follows the correct temporal evidence trace during reasoning, rather than merely producing the correct final answer.

\noindent
\textbf{QA and Training Annotation.}
We generate QA pairs directly from the program metadata using task-specific templates, ensuring that each question is paired with an exact, automatically computed answer. For the training split, we further generate reasoning traces by conditioning proprietary models (i.e.,GPT-5.4 ) on the video, metadata, question, and answer. We then manually curate ten reasoning templates for each task category to keep the generated reasoning aligned with the underlying temporal evidence trace while preserving controlled linguistic variation.

\noindent
\textbf{Metric Design.} Since \textsc{Video-MME-Logical} provides program-derived ground-truth answers and states, we use exact-match accuracy as the main evaluation metric. \textit{Multiple-choice questions.} We extract the selected option from the unified answer tag and count it as correct only if it exactly matches the ground-truth option. \textit{Fill-in questions.} We extract the tagged answer, and count it as correct only if it exactly matches the ground-truth answer. \textit{Intermediate-state questions.} In \textsc{Video-MME-Logical-S}, each example additionally contains program-recorded intermediate states for process verification. Models are required to output structured intermediate information in the same answer tag. We canonicalize the predicted structure and count it as correct only if it exactly matches the corresponding program-recorded state. Overall accuracy is computed as $\mathrm{Acc}=N_{\mathrm{correct}}/N_{\mathrm{total}}$.

\section{Experiments}
We study \textsc{Video-MME-Logical} from three aspects: final-answer evaluation, intermediate-state evaluation, and supervised fine-tuning scaling.

\subsection{Implementation Details}

\begin{table*}[t]
  \centering
  \scriptsize
  \setlength{\tabcolsep}{2pt}
  \resizebox{\textwidth}{!}{%
  \begin{tabular}{lccccccccccccccccccc}
    \toprule
    Models & Overall
    & \multicolumn{3}{c}{Avg.}
    & \multicolumn{3}{c}{State.}
    & \multicolumn{3}{c}{Count.}
    & \multicolumn{3}{c}{Order.}
    & \multicolumn{3}{c}{Spat.}
    & \multicolumn{3}{c}{Struct.} \\
    \cmidrule(lr){3-5}
    \cmidrule(lr){6-8}
    \cmidrule(lr){9-11}
    \cmidrule(lr){12-14}
    \cmidrule(lr){15-17}
    \cmidrule(lr){18-20}
    & & E & M & H & E & M & H & E & M & H & E & M & H & E & M & H & E & M & H \\
    \midrule
    Human Level & 95.9 & 98.4 & 95.9 & 93.4 & 99.2 & 96.5 & 93.5 & 97.6 & 95.2 & 93.2 & 99.0 & 95.5 & 93.5 & 98.5 & 97.0 & 93.5 & 97.0 & 95.0 & 93.5 \\
    \midrule
    \rowcolor{tablegroupblue}
    \multicolumn{20}{l}{\textbf{Open-source Instruct Models}} \\
    Qwen3-VL-8B-Instruct & 11.9 & 13.4 & 12.8 & 9.6 & 11.0 & 8.8 & 4.8 & 9.6 & 0.0 & 0.4 & 18.0 & 20.5 & 19.5 & 12.0 & 14.5 & 12.5 & 16.5 & 20.0 & 11.0 \\
    Qwen3-VL-30B-A3B-Instruct & 11.8 & 14.5 & 12.4 & 8.7 & 11.8 & 8.8 & \textbf{4.8} & 11.6 & 0.4 & 0.0 & 17.0 & 20.5 & 14.0 & 18.5 & 20.5 & 12.5 & 13.5 & 12.0 & 12.0 \\
    Qwen3-Omni-30B-A3B-Instruct & 5.8 & 6.3 & 6.1 & 4.9 & 3.8 & 4.0 & 1.0 & 2.8 & 0.4 & 0.0 & 6.5 & 9.0 & 10.5 & 9.5 & 6.5 & 3.5 & 9.0 & 10.5 & 9.5 \\
    Qwen2.5-VL-3B-Instruct & 1.9 & 3.1 & 1.5 & 1.3 & 1.0 & 1.0 & 0.0 & 4.8 & 0.8 & 0.0 & 0.0 & 0.5 & 0.5 & 0.0 & 0.0 & 1.5 & 9.5 & 5.0 & 4.5 \\
    Qwen2.5-VL-7B-Instruct & 7.4 & 10.3 & 7.7 & 4.3 & 9.5 & 6.8 & 2.2 & 7.2 & 0.0 & 0.4 & 0.0 & 0.0 & 0.0 & 23.0 & 26.0 & 10.5 & 12.0 & 5.5 & 8.5 \\
    Qwen2.5-VL-72B-Instruct & 12.5 & 15.2 & 13.1 & 9.1 & 7.8 & 5.0 & 4.0 & 10.8 & 1.6 & 0.0 & 19.5 & 20.5 & 15.5 & 19.0 & 22.5 & 12.5 & 19.0 & 16.0 & 13.5 \\
    InternVL3.5-8B-Instruct & 12.1 & 13.8 & 13.5 & 8.9 & 7.2 & 4.5 & 2.2 & 8.0 & 4.0 & 0.8 & 16.5 & 18.5 & 12.0 & 18.0 & 20.0 & 16.0 & 19.5 & 20.5 & 13.5 \\
    InternVL3.5-30B-A3B-Instruct & 8.7 & 9.5 & 9.7 & 7.0 & 9.8 & 7.5 & 2.5 & 5.2 & 4.4 & 3.6 & 3.5 & 1.0 & 3.0 & 15.0 & 18.5 & 14.0 & 14.0 & 17.0 & 12.0 \\
    LLaVA-Video-7B-Qwen2 & 0.0 & 0.1 & 0.0 & 0.0 & 0.2 & 0.0 & 0.0 & 0.0 & 0.0 & 0.0 & 0.0 & 0.0 & 0.0 & 0.0 & 0.0 & 0.0 & 0.0 & 0.0 & 0.0 \\
    LLaVA-Video-72B-Qwen2 & 2.4 & 4.7 & 1.9 & 0.7 & 4.8 & 5.2 & 2.8 & 13.2 & 1.2 & 0.8 & 0.0 & 0.0 & 0.0 & 5.5 & 3.0 & 0.0 & 0.0 & 0.0 & 0.0 \\
    KimiVL-16B-A3B-Instruct & 2.9 & 5.4 & 2.2 & 0.9 & 7.0 & 4.8 & 1.5 & 11.6 & 0.0 & 0.0 & 2.5 & 3.0 & 1.5 & 1.0 & 0.0 & 0.0 & 5.0 & 3.5 & 1.5 \\
    \midrule
    \rowcolor{tablegroupblue}
    \multicolumn{20}{l}{\textbf{Open-source Thinking Models}} \\
    Qwen3-VL-8B-Think & 6.6 & 7.9 & 5.8 & 6.0 & 0.8 & 0.8 & 1.8 & 4.4 & 3.6 & 1.6 & 5.5 & 4.0 & 6.0 & 19.5 & 15.5 & 13.0 & 9.5 & 5.0 & 7.5 \\
    Qwen3-VL-30B-A3B-Think & 10.3 & 16.0 & 8.7 & 6.1 & 4.0 & 4.8 & 2.8 & 21.6 & 7.2 & 3.6 & 28.5 & 23.0 & 20.5 & 16.0 & 5.5 & 1.5 & 10.0 & 3.0 & 2.0 \\
    Qwen3-Omni-30B-A3B-Think & 6.2 & 6.6 & 5.2 & 6.8 & 1.8 & 1.2 & 4.2 & 2.8 & 2.0 & 1.2 & 9.0 & 13.0 & 15.5 & 8.5 & 7.0 & 3.5 & 11.0 & 2.5 & 9.5 \\
    KimiVL-16B-A3B-Think & 7.6 & 10.2 & 6.8 & 5.8 & 7.0 & 5.0 & 3.8 & 4.0 & 0.4 & 0.0 & 5.0 & 5.0 & 5.0 & 20.0 & 13.5 & 10.0 & 15.0 & 10.0 & 10.0 \\
    \midrule
    \rowcolor{tablegroupblue}
    \multicolumn{20}{l}{\textbf{Proprietary Models}} \\
    GPT-5.4 & 22.7 & 31.7 & 20.3 & 16.1 & 8.8 & 4.5 & 3.8 & 56.4 & 16.4 & 5.6 & 38.0 & 37.0 & 32.0 & 29.0 & 22.5 & 24.0 & 26.5 & 21.0 & 15.0 \\
    Gemini-3.1 Pro & \textbf{28.6} & \textbf{33.1} & \textbf{24.1} & \textbf{20.6} & \textbf{14.0} & \textbf{9.8} & 3.0 & \textbf{58.4} & \textbf{36.0} & \textbf{30.8} & \textbf{38.5} & \textbf{34.0} & \textbf{32.5} & \textbf{35.5} & \textbf{27.5} & \textbf{35.5} & \textbf{32.0} & \textbf{24.5} & \textbf{16.5} \\
    \bottomrule
  \end{tabular}%
  }
  \caption{Main results on \textsc{Video-MME-Logical}. E/M/H denote easy, medium, and hard settings; State., Count., Order., Spat., and Struct. denote the five reasoning dimensions.}
  \label{tab:main-results}
\end{table*}

\noindent
\textbf{Benchmark Models.} We evaluate a diverse set of video-capable MLLMs on \textsc{Video-MME-Logical}, grouped into three categories according to model type. First, \textit{open-source instruct models} include Qwen2.5-VL-3B/7B/72B~\citep{qwen25vl}, Qwen3-VL and Qwen3-Omni variants~\citep{qwen3vl}, InternVL3.5 variants~\citep{internvl35}, LLaVA-Video-7B/72B-Qwen2~\citep{llavavideo}, and KimiVL-16B-A3B-Instruct~\citep{kimivl}. Second, \textit{open-source thinking models} include Qwen3-VL, Qwen3-Omni, and KimiVL thinking variants, which are designed to produce explicit reasoning traces before answering. Third, \textit{proprietary models} include GPT-5.4~\citep{openai2026gpt54} and gemini-3.1 Pro~\citep{google2025gemini3pro}. All evaluations are conducted under zero-shot settings, with videos sampled at 2 FPS.

\noindent
\textbf{Human-Level Performance.} We further estimate human performance on a sampled subset of \textsc{Video-MME-Logical}. Human evaluators independently answer each question under the same visual input setting as models, and their predictions are evaluated using the same metrics. This provides an approximate reference point for interpreting model performance and assessing whether the benchmark remains solvable for human annotators.

\noindent
\subsection{Final-Answer Evaluation}

Table~\ref{tab:main-results} reports final-answer accuracy over all 25 task categories and three difficulty levels. Overall, \textsc{Video-MME-Logical} remains highly challenging for current MLLMs. Human performance reaches 95.9\% overall accuracy, whereas the strongest evaluated model, gemini-3.1 Pro, achieves only 28.6\%. These results show that a large gap still remains between human performance and current models.

\noindent
\textbf{Thinking does not necessarily improve temporal-logical reasoning.}
Although KimiVL-16B-A3B improves from 2.9\% in the instruct setting to 7.6\% in the thinking setting, this trend is not consistent across model families. Qwen3-VL-8B drops from 11.9\% to 6.6\% when switching from instruct to thinking, and Qwen3-VL-30B-A3B drops from 11.8\% to 10.3\%. This suggests that generating a reasoning trace is not sufficient by itself; the trace must be grounded in the correct visual evidence. 

\noindent
\textbf{Controlled difficulty exposes different degradation patterns.} As difficulty increases, GPT-5.4 drops from 31.7\% on easy tasks to 16.1\% on hard tasks, a degradation of 15.6\%, while gemini-3.1 Pro drops from 33.1\% to 20.6\%, a degradation of 12.5\%. The smaller degradation of gemini-3.1 Pro suggests stronger robustness on harder examples. However, its hard-task performance still remains far below human level, indicating that longer temporal horizons and higher reasoning complexity remain challenging for current models.

\begin{table*}[!t]
  \centering
  \scriptsize
  \setlength{\tabcolsep}{3pt}
  \resizebox{\textwidth}{!}{%
  \begin{tabular}{lccccccccccccc}
    \toprule
    Models & Overall
    & \multicolumn{3}{c}{Avg.}
    & \multicolumn{3}{c}{State.-S}
    & \multicolumn{3}{c}{Count.-S}
    & \multicolumn{3}{c}{Order.-S} \\
    \cmidrule(lr){3-5}
    \cmidrule(lr){6-8}
    \cmidrule(lr){9-11}
    \cmidrule(lr){12-14}
    & & E & M & H & E & M & H & E & M & H & E & M & H \\
    \midrule
    Human Level & 96.1 & 98.5 & 96.0 & 93.8 & 99.0 & 96.5 & 93.5 & 97.0 & 95.0 & 94.0 & 99.0 & 96.0 & 94.0 \\
    \midrule
    \rowcolor{tablegroupblue}
    \multicolumn{14}{l}{\textbf{Open-source Instruct Models}} \\
    Qwen3-VL-8B-Instruct & 0.0 & 0.0 & 0.0 & 0.0 & 0.0 & 0.0 & 0.0 & 0.0 & 0.0 & 0.0 & 0.0 & 0.0 & 0.0 \\
    Qwen3-VL-30B-A3B-Instruct & 0.1 & 0.3 & 0.0 & 0.0 & 0.0 & 0.0 & 0.0 & 1.0 & 0.0 & 0.0 & 0.0 & 0.0 & 0.0 \\
    Qwen3-Omni-30B-A3B-Instruct & 0.0 & 0.0 & 0.0 & 0.0 & 0.0 & 0.0 & 0.0 & 0.0 & 0.0 & 0.0 & 0.0 & 0.0 & 0.0 \\
    Qwen2.5-VL-3B-Instruct & 0.0 & 0.0 & 0.0 & 0.0 & 0.0 & 0.0 & 0.0 & 0.0 & 0.0 & 0.0 & 0.0 & 0.0 & 0.0 \\
    Qwen2.5-VL-7B-Instruct & 0.1 & 0.2 & 0.0 & 0.0 & 0.5 & 0.0 & 0.0 & 0.0 & 0.0 & 0.0 & 0.0 & 0.0 & 0.0 \\
    Qwen2.5-VL-72B-Instruct & 0.1 & 0.2 & 0.0 & 0.0 & 0.5 & 0.0 & 0.0 & 0.0 & 0.0 & 0.0 & 0.0 & 0.0 & 0.0 \\
    InternVL3.5-8B-Instruct & 0.0 & 0.0 & 0.0 & 0.0 & 0.0 & 0.0 & 0.0 & 0.0 & 0.0 & 0.0 & 0.0 & 0.0 & 0.0 \\
    InternVL3.5-30B-A3B-Instruct & 0.1 & 0.3 & 0.0 & 0.0 & 1.0 & 0.0 & 0.0 & 0.0 & 0.0 & 0.0 & 0.0 & 0.0 & 0.0 \\
    LLaVA-Video-7B-Qwen2 & 0.1 & 0.2 & 0.0 & 0.0 & 0.5 & 0.0 & 0.0 & 0.0 & 0.0 & 0.0 & 0.0 & 0.0 & 0.0 \\
    LLaVA-Video-72B-Qwen2 & 0.0 & 0.0 & 0.0 & 0.0 & 0.0 & 0.0 & 0.0 & 0.0 & 0.0 & 0.0 & 0.0 & 0.0 & 0.0 \\
    KimiVL-16B-A3B-Instruct & 0.1 & 0.2 & 0.0 & 0.0 & 0.5 & 0.0 & 0.0 & 0.0 & 0.0 & 0.0 & 0.0 & 0.0 & 0.0 \\
    \midrule
    \rowcolor{tablegroupblue}
    \multicolumn{14}{l}{\textbf{Open-source Thinking Models}} \\
    Qwen3-VL-8B-Think & 0.6 & 1.3 & 0.3 & 0.0 & 0.0 & 0.0 & 0.0 & 1.0 & 0.0 & 0.0 & 3.0 & 1.0 & 0.0 \\
    Qwen3-VL-30B-A3B-Think & 3.6 & 9.0 & 1.3 & 0.3 & 0.0 & 0.0 & 0.0 & 11.0 & 0.0 & 0.0 & 16.0 & 4.0 & 1.0 \\
    Qwen3-Omni-30B-A3B-Think & 1.2 & 1.7 & 1.3 & 0.7 & 0.0 & 0.0 & 0.0 & 0.0 & 0.0 & 0.0 & 5.0 & 4.0 & 2.0 \\
    KimiVL-16B-A3B-Think & 0.0 & 0.0 & 0.0 & 0.0 & 0.0 & 0.0 & 0.0 & 0.0 & 0.0 & 0.0 & 0.0 & 0.0 & 0.0 \\
    \midrule
    \rowcolor{tablegroupblue}
    \multicolumn{14}{l}{\textbf{Proprietary Models}} \\
    GPT-5.4 & \textbf{17.4} & \textbf{30.8} & \textbf{13.7} & \textbf{7.7} & 2.5 & 0.0 & 0.0 & \textbf{63.0} & \textbf{20.0} & 9.0 & \textbf{27.0} & \textbf{21.0} & \textbf{14.0} \\
    Gemini-3.1 Pro & 10.8 & 18.7 & 8.5 & 5.2 & \textbf{8.0} & \textbf{3.5} & \textbf{0.5} & 35.0 & 12.0 & \textbf{10.0} & 13.0 & 10.0 & 5.0 \\
    \bottomrule
  \end{tabular}%
  }
  \caption{Main results on \textsc{Video-MME-Logical-S}.}
  \label{tab:step-subset-results}
\end{table*}

\begin{table*}[!t]
  \centering
  \scriptsize
  \setlength{\tabcolsep}{2pt}
  \resizebox{\textwidth}{!}{%
  \begin{tabular}{lccccccccccccccccccc}
    \toprule
    Models & Overall
    & \multicolumn{3}{c}{Avg.}
    & \multicolumn{3}{c}{State.}
    & \multicolumn{3}{c}{Count.}
    & \multicolumn{3}{c}{Order.}
    & \multicolumn{3}{c}{Spat.}
    & \multicolumn{3}{c}{Struct.} \\
    \cmidrule(lr){3-5}
    \cmidrule(lr){6-8}
    \cmidrule(lr){9-11}
    \cmidrule(lr){12-14}
    \cmidrule(lr){15-17}
    \cmidrule(lr){18-20}
    & & E & M & H & E & M & H & E & M & H & E & M & H & E & M & H & E & M & H \\
    \midrule
    \rowcolor{tablegroupblue}
    \multicolumn{20}{l}{\textbf{Base Model}} \\
    Qwen3-VL-8B-Instruct & 11.9 & 13.4 & 12.8 & 9.6 & 11.0 & 8.8 & 4.8 & 9.6 & 0.0 & 0.4 & 18.0 & 20.5 & 19.5 & 12.0 & 14.5 & 12.5 & 16.5 & 20.0 & 11.0 \\
    \midrule
    \rowcolor{tablegroupblue}
    \multicolumn{20}{l}{\textbf{SFT Scaling -- Thinking}} \\
    Ours-25K-Thinking & 36.8 & 46.7 & \textbf{35.2} & \textbf{28.4} & 37.8 & \textbf{13.8} & 6.5 & \textbf{43.6} & \textbf{28.4} & \textbf{20.4} & 50.5 & 43.5 & 34.5 & 60.0 & 59.0 & 53.0 & 41.5 & 31.5 & 27.5 \\
    Ours-125K-Thinking & 36.9 & 52.5 & 32.6 & 25.6 & 39.0 & 9.8 & 4.2 & 42.4 & 12.0 & 1.6 & 54.5 & 43.0 & 33.0 & 75.0 & 66.0 & 60.0 & \textbf{51.5} & 32.5 & 29.0 \\
    Ours-250K-Thinking & 36.9 & 53.6 & 33.2 & 24.0 & \textbf{40.8} & 7.5 & 3.0 & 43.2 & 8.4 & 1.6 & 57.5 & \textbf{51.0} & 31.5 & 83.0 & 62.0 & 56.0 & 43.5 & 37.0 & 28.0 \\
    Ours-375K-Thinking & \textbf{39.2} & \textbf{54.8} & 34.7 & 28.1 & 40.2 & 7.5 & 4.2 & 41.6 & 6.0 & 3.2 & \textbf{60.5} & 49.0 & \textbf{37.0} & 82.5 & \textbf{73.0} & \textbf{69.5} & 49.0 & \textbf{38.0} & 26.5 \\
    Ours-500K-Thinking & 37.7 & 52.0 & 34.0 & 27.0 & 33.5 & 6.0 & 3.5 & 40.0 & 9.6 & 2.4 & 57.0 & 48.0 & 33.0 & \textbf{87.0} & 71.0 & 66.0 & 42.5 & 35.5 & \textbf{30.0} \\
    \midrule
    \rowcolor{tablegroupblue}
    \multicolumn{20}{l}{\textbf{SFT Scaling -- Instruct}} \\
    Ours-25K-Instruct & 23.9 & 33.2 & 21.2 & 17.3 & 23.5 & 7.0 & 4.5 & 32.0 & 4.4 & 0.8 & 19.5 & 10.0 & 6.0 & 60.5 & 51.5 & 48.0 & 30.5 & 33.0 & 27.0 \\
    Ours-125K-Instruct & 28.9 & 40.6 & 26.7 & 19.4 & 24.8 & 6.8 & 5.0 & 29.2 & 4.8 & 1.2 & 43.0 & 30.5 & 13.5 & 74.5 & 63.5 & 55.5 & 31.5 & 28.0 & 22.0 \\
    Ours-250K-Instruct & 30.4 & 41.5 & 26.9 & 22.6 & 28.8 & 10.8 & \textbf{7.2} & 30.0 & 1.6 & 0.0 & 44.5 & 37.5 & 30.5 & 75.0 & 63.0 & 55.5 & 29.5 & 21.5 & 20.0 \\
    Ours-375K-Instruct & 30.6 & 41.9 & 27.3 & 22.6 & 27.5 & 6.8 & 6.8 & 30.4 & 2.8 & 0.4 & 45.0 & 35.5 & 29.5 & 77.5 & 69.0 & 57.0 & 29.0 & 22.5 & 19.5 \\
    Ours-500K-Instruct & 27.3 & 38.5 & 25.1 & 18.2 & 27.2 & 9.2 & 6.5 & 32.0 & 4.0 & 1.2 & 22.5 & 18.0 & 3.0 & 80.5 & 69.5 & 54.5 & 30.5 & 25.0 & 26.0 \\
    \bottomrule
  \end{tabular}%
  }
\caption{SFT scaling results on \textsc{Video-MME-Logical}.}·
  \label{tab:sft-scaling}
\end{table*}

\subsection{Intermediate-State Evaluation}

\noindent\Cref{tab:step-subset-results} evaluates \textsc{Video-MME-Logical-S}. This subset tests whether models can generate accurate intermediate logical reasoning rather than only produce final answers.

\noindent
\textbf{Final-answer accuracy can hide intermediate-state failures.}
Although gemini-3.1 Pro outperforms GPT-5.4 on easy final-answer State, Count., and Order. tasks (14.0\% vs. 8.8\%, 58.4\% vs. 56.4\%, and 38.5\% vs. 38.0\%), GPT-5.4 is stronger on the easy step-subset categories: 63.0\% vs. 35.0\% on Count.-S and 27.0\% vs. 13.0\% on Order.-S. This mismatch shows that a higher final-answer score does not necessarily imply that the model can generate the process for the final answer.

\noindent
\textbf{Proprietary models lead on intermediate-state evaluation.}
GPT-5.4 and gemini-3.1 Pro achieve 17.4\% and 10.8\% on the intermediate-state evaluation subset, respectively, while the strongest open-source model, Qwen3-VL-30B-A3B-Think, reaches only 3.6\%. GPT-5.4 is therefore 4.8$\times$ higher than the strongest open-source model, and gemini-3.1 Pro is 3.0$\times$ higher. This large gap highlights the value of \textsc{Video-MME-Logical-S}.

\begin{figure*}[t]

  \centering
  \includegraphics[width=\textwidth]{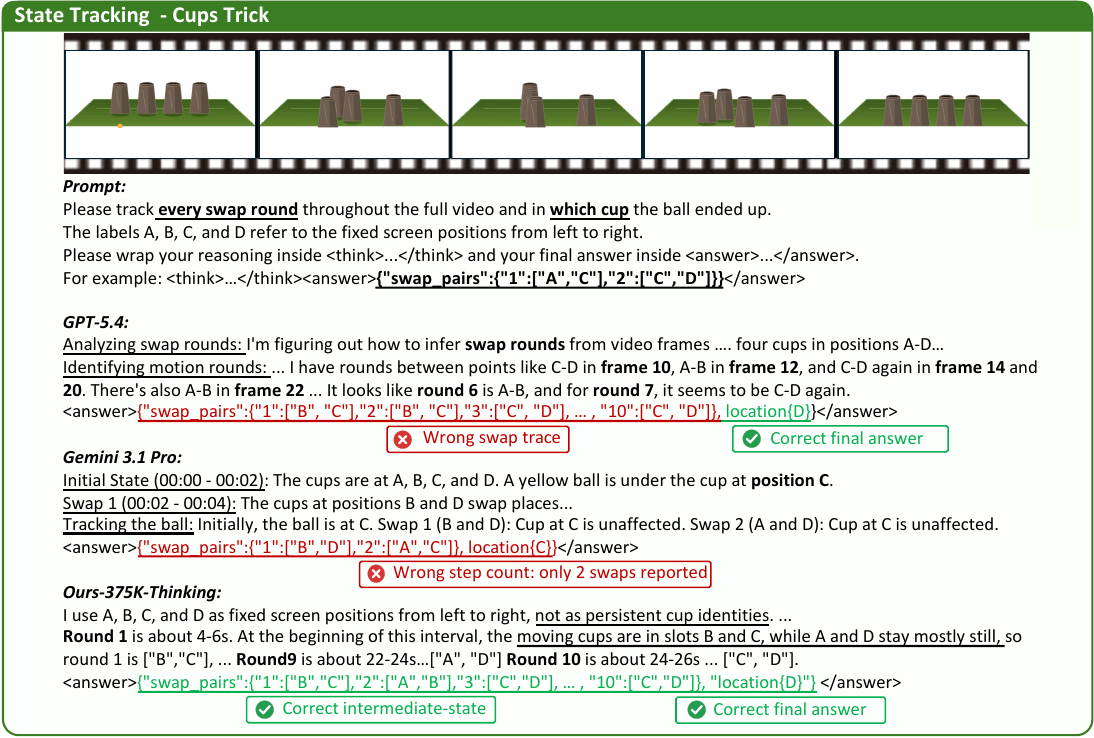}
  \caption{A qualitative example of intermediate-state evaluation on a state-tracking task.}
  \label{fig:quality-example}

\end{figure*}

\subsection{SFT Scaling Analysis}

\Cref{tab:sft-scaling} studies SFT using Qwen3-VL-8B as the base model. We sample training data from the 500K training split at five sizes: 25K, 125K, 250K, 375K, and 500K, with balanced proportions across task categories. \textit{Ours-*K-Instruct} denotes models trained with answer supervision, while \textit{Ours-*K-Thinking} denotes models trained with reasoning trajectories. Since the training data is constructed from easy-level instances, evaluation on medium and hard settings further tests whether the learned temporal-logical reasoning behaviors can generalize to longer duration and higher complexity. We provide training details in Appendix~\ref{appendix: training details}.

\noindent
\textbf{Data scaling brings clear but saturating gains.}
Increasing the training size to 375K improves overall accuracy to 39.2\%, the best result in the scaling table. However, using the full 500K training set reduces performance to 37.7\%. This trend suggests that the current model and SFT recipe can learn useful temporal-logical reasoning behaviors, but simply adding more supervised data does not bring sustained improvement.

\noindent
\textbf{Generalization to harder settings remains limited.}
Ours-375K-Thinking reaches 54.8\% Avg. E, indicating that the model learns transferable temporal-logical behaviors from easy-level training data. However, medium and hard performance do not improve consistently with scale: Ours-25K-Thinking is slightly higher than Ours-375K-Thinking by 0.5\% on Avg. M and 0.3\% on Avg. H. These results suggest that easy-level supervision can generalize to some more complex settings, but it does not provide stable gains under longer temporal horizons and higher reasoning complexity.

\noindent\textbf{Visualization Analysis.}
Figure~\ref{fig:quality-example} illustrates the role of \textsc{Video-MME-Logical-S} in diagnosing intermediate-state reasoning, using state tracking as an example. GPT-5.4 predicts the correct final location but produces an incorrect swap trace, showing that final-answer accuracy can hide process-level errors. Gemini 3.1 Pro reports an incorrect number of reasoning steps, indicating incomplete temporal state updates. In contrast, Ours-375K-Thinking recovers both the intermediate swap sequence and the final answer. This demonstrates that verifiable intermediate states can distinguish genuine temporal-logical reasoning from superficially correct final answers but flawed intermediate reasoning.

\section{Conclusion}

We introduced \textsc{VIDEO-MME-LOGICAL}, a controlled diagnostic benchmark for video temporal-logical reasoning. By organizing 25 tasks around five temporal-logical operations and providing difficulty-controlled settings with intermediate-state diagnostics, our benchmark isolates whether MLLMs can maintain, update, and compose visual evidence over time. Experiments reveal a substantial human-model gap, especially under longer temporal horizons, higher reasoning complexity, and process-level evaluation. Our SFT scaling study further shows that supervised data improves performance but quickly saturates, suggesting that naive supervised scaling in our setting is insufficient for robust temporal-logical reasoning. We hope \textsc{VIDEO-MME-LOGICAL} will support more precise diagnosis of video reasoning failures and encourage future models with stronger temporal-logical reasoning capabilities.

\section*{Limitations}
This work has several limitations. First, \textsc{Video-MME-Logical} is built from procedurally generated videos. This design enables scalable data generation, controllable difficulty, and verifiable intermediate states, but it also introduces a gap from natural videos in visual appearance, scene diversity, and real-world ambiguity. However, natural videos are difficult to annotate at a large scale with reliable temporal states, exact answers, and process-level supervision, which motivates our controlled diagnostic setting. Second, our scaling experiments are conducted with an 8B MLLM. Larger models, such as 72B-scale MLLMs, may exhibit different scaling behaviors, especially in long-horizon state maintenance and compositional reasoning, and we leave a broader model-scale study to future work. Third, our intermediate-state evaluation relies on task-specific structured outputs and exact-match scoring. While this makes the evaluation reproducible and directly tied to program-derived ground truth, it may penalize semantically valid reasoning traces that use different surface forms or alternative but equivalent descriptions.

\bibliography{references}

\clearpage

\appendix
\onecolumn

\section{Benchmark Details and Task Taxonomy}
\label{app:task-details}

\begin{center}
\begingroup
  \scriptsize
  \setlength{\tabcolsep}{4pt}
  \renewcommand{\arraystretch}{0.92}
  \begin{tabular}{p{0.29\textwidth}p{0.11\textwidth}p{0.48\textwidth}}
    \toprule
    Task & Format & Description \\
    \midrule
    \multicolumn{3}{l}{\textbf{\textcolor[HTML]{7CCB4D}{State Tracking (8)}}} \\
    Cup Trick & MC & Locate hidden ball after swaps \\
    Cup Trick-S & Fill-in & Recover cup swap sequence \\
    Cup Shuffle & MC & Track empty cup through shuffles \\
    Cup Shuffle-S & Fill-in & Recover three-cup shuffle trace \\
    Card Relocation & MC & Locate target card after moves \\
    Card Relocation-S & Fill-in & Recover card position history \\
    Card Shuffle & Fill-in & Count target card relocations \\
    Card Shuffle-S & Fill-in & Recover card move sequence \\
    \midrule
    \multicolumn{3}{l}{\textbf{\textcolor[HTML]{FF8B4A}{Structural Composition (4)}}} \\
    Falling Shape Count & Fill-in & Count falling target shapes \\
    3D Maze Route & MC & Match route through 3D maze \\
    Occlusion Object Count & Fill-in & Count objects under occlusion \\
    Hidden Container Inference & MC & Infer hidden container shape \\
    \midrule
    \multicolumn{3}{l}{\textbf{\textcolor[HTML]{36BDF2}{Dynamic Spatiality (4)}}} \\
    Maze Trace & Fill-in & Count turns along route \\
    Rotation Center & MC & Locate image rotation center \\
    Trajectory Intersection & MC & Count trajectory intersections \\
    Speed Comparison & MC & Compare object motion speeds \\
    \midrule
    \multicolumn{3}{l}{\textbf{\textcolor[HTML]{EA7AD8}{Temporal Ordering (4)}}} \\
    Keyboard Sequence & Fill-in & Read ordered letter sequence \\
    Keyboard Sequence-S & Fill-in & Recover letter reveal order \\
    Neon Word & MC & Identify word from sequential flashes \\
    Neon Word-Step & Fill-in & Recover word formation sequence \\
    \midrule
    \multicolumn{3}{l}{\textbf{\textcolor[HTML]{FF6FB5}{Sequential Counting (5)}}} \\
    Symbol & Fill-in & Count matching symbols over time \\
    Symbol-S & Fill-in & Recover symbol reveal sequence \\
    Cube Structure Count & Fill-in & Count cubes in 3D structure \\
    Grid Activation & Fill-in & Count unique activated cells \\
    Grid Activation-S & Fill-in & Recover cell activation trace \\
    \bottomrule
  \end{tabular}
  \captionof{table}{Task categories grouped by cognitive category. Format denotes the answer type: MC for multiple-choice and Fill-in for open-ended numeric, string, or JSON answers.}
  \label{tab:task-taxonomy}
\endgroup
\end{center}

\begin{figure}[h]
  \centering
  \includegraphics[width=0.96\textwidth,height=0.82\textheight,keepaspectratio]{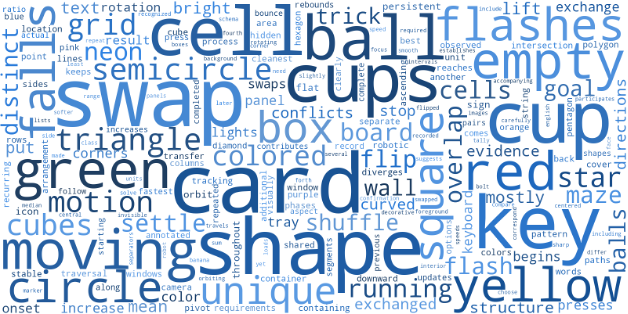}
  \caption{Word cloud of \textsc{Video-MME-Logical}.}
  \label{fig:word-cloud}
\end{figure}

\section{Training Configuration for the Qwen3-VL-8B SFT Experiments}
\label{appendix: training details}
\begin{table}[t]
\centering
\label{tab:sft-training-config}
\small
\begin{tabular}{ll}
\toprule
\textbf{Item} & \textbf{Value} \\
\midrule
Base model & Qwen3-VL-8B-Instruct \\
Training stage & Supervised fine-tuning (SFT) \\
Fine-tuning type & Full-parameter fine-tuning \\
Frozen modules & Vision tower; multimodal projector \\
DeepSpeed & ZeRO-2 \\
Flash attention & FlashAttention-2 \\
Template & Qwen3-VL no-thinking \\
Dataset mixture & Video-only logic SFT + image logic SFT \\
Maximum samples & 1,000,000 \\
Cutoff length & 16,384 tokens \\
Image max pixels & 802,816 \\
Video max pixels & 100,352 \\
Maximum video length & 128 frames \\
Video sampling rate & 2 FPS \\
Epochs & 2 \\
Batch size per device & 4 \\
Gradient accumulation & 1 \\
Learning rate & $2.0 \times 10^{-5}$ \\
Weight decay & $1.0 \times 10^{-6}$ \\
Scheduler & Cosine \\
Warmup ratio & 0.1 \\
Precision & BF16 \\
Save interval & 1,000 steps \\
Checkpoint limit & 5 \\

GPU hours & H800 x 8 x 48h \\
\bottomrule
\end{tabular}
\caption{Training configuration for the Qwen3-VL-8B SFT experiments.}
\end{table}

\newpage

\section{Human Evaluation Protocol}
To estimate human-level performance, we sampled 3750 examples from \textsc{Video-MME-Logical}. The sampled examples were evaluated by three human annotators under the same visual-input setting used for model evaluation: annotators were shown the video and the corresponding question, but were not given access to program metadata, ground-truth answers, intermediate-state annotations, or model predictions. The annotators were instructed to answer each question according to the same output format used in our automatic evaluation, including structured answers for intermediate-state tasks when required. Human responses were then scored with the same evaluation script used for model outputs. Annotators were compensated at a rate of 50 USD per hour. The resulting human-level score is used only as a reference for benchmark solvability and for contextualizing the gap between human performance and current MLLMs.

\section{More Visualization Analysis}

\begin{figure*}[h]
  \centering
  \includegraphics[width=0.96\textwidth,height=0.82\textheight,keepaspectratio]{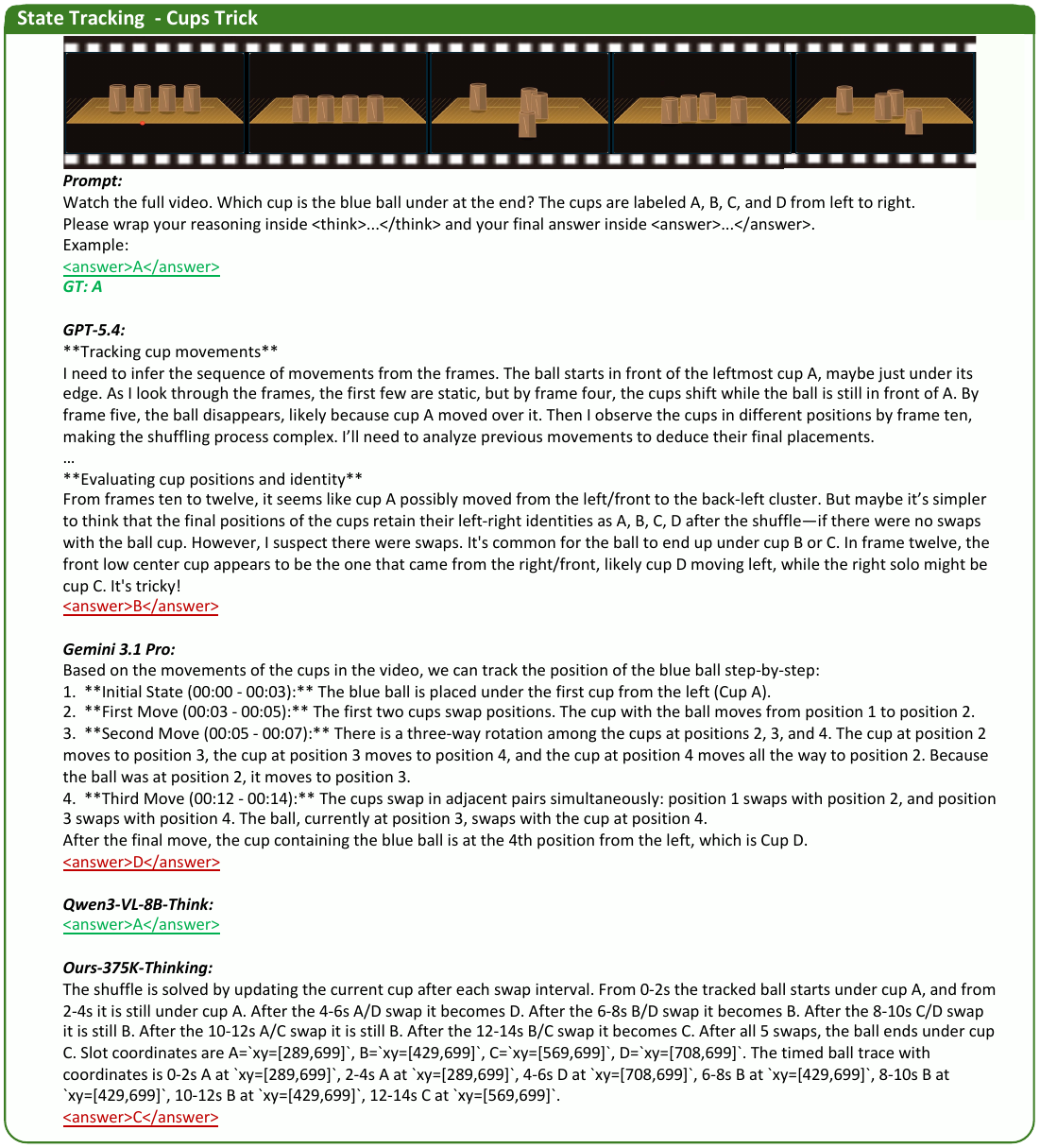}
  \caption{Additional visual examples from \textsc{Video-MME-Logical}.}
  \label{fig:appendix-show-data-01}
\end{figure*}

\begin{figure*}[h]
  \centering
  \includegraphics[width=0.96\textwidth,height=0.82\textheight,keepaspectratio]{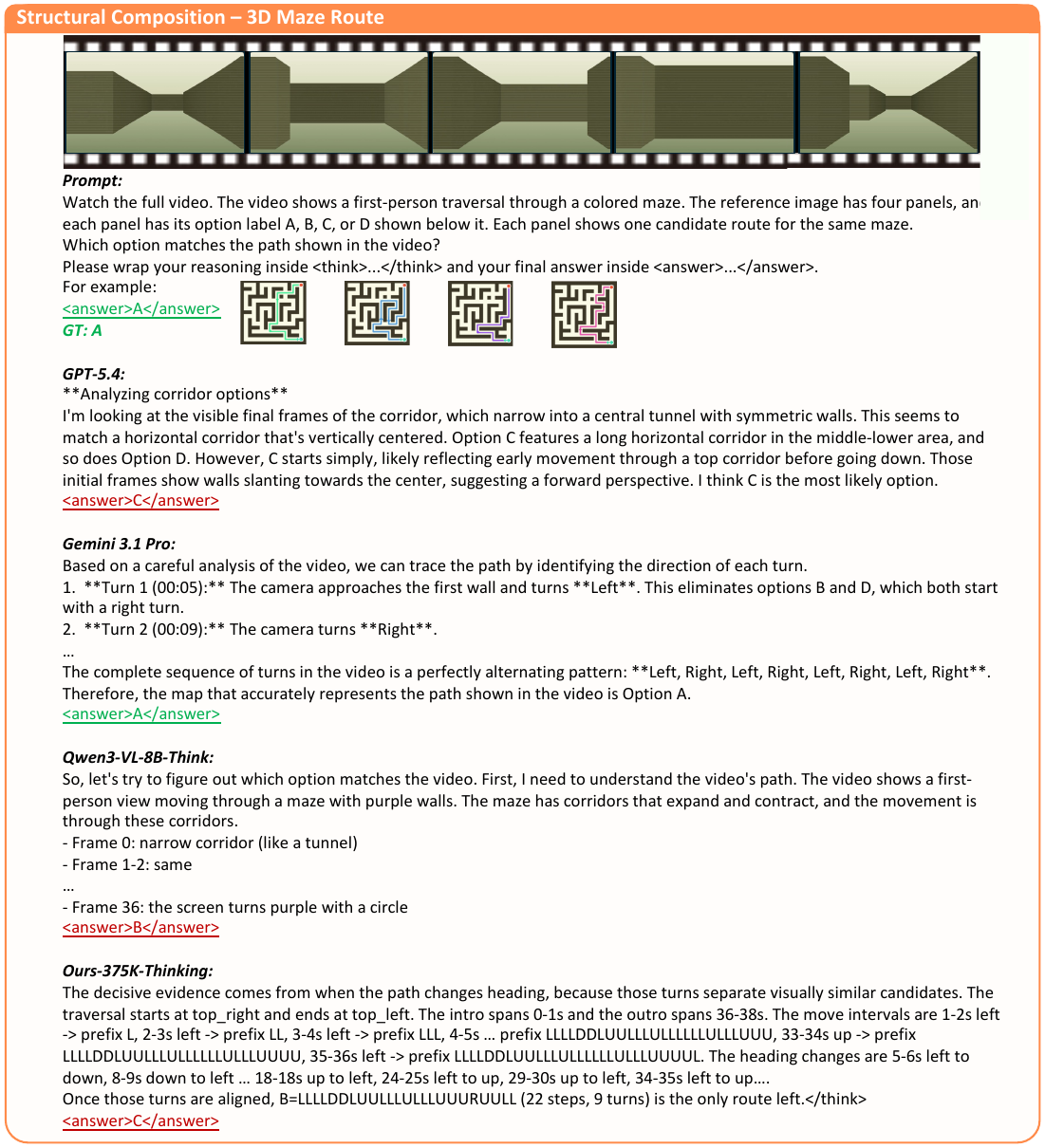}
  \caption{Additional visual examples from \textsc{Video-MME-Logical}.}
  \label{fig:appendix-show-data-02}
\end{figure*}

\begin{figure*}[h]
  \centering
  \includegraphics[width=0.96\textwidth,height=0.82\textheight,keepaspectratio]{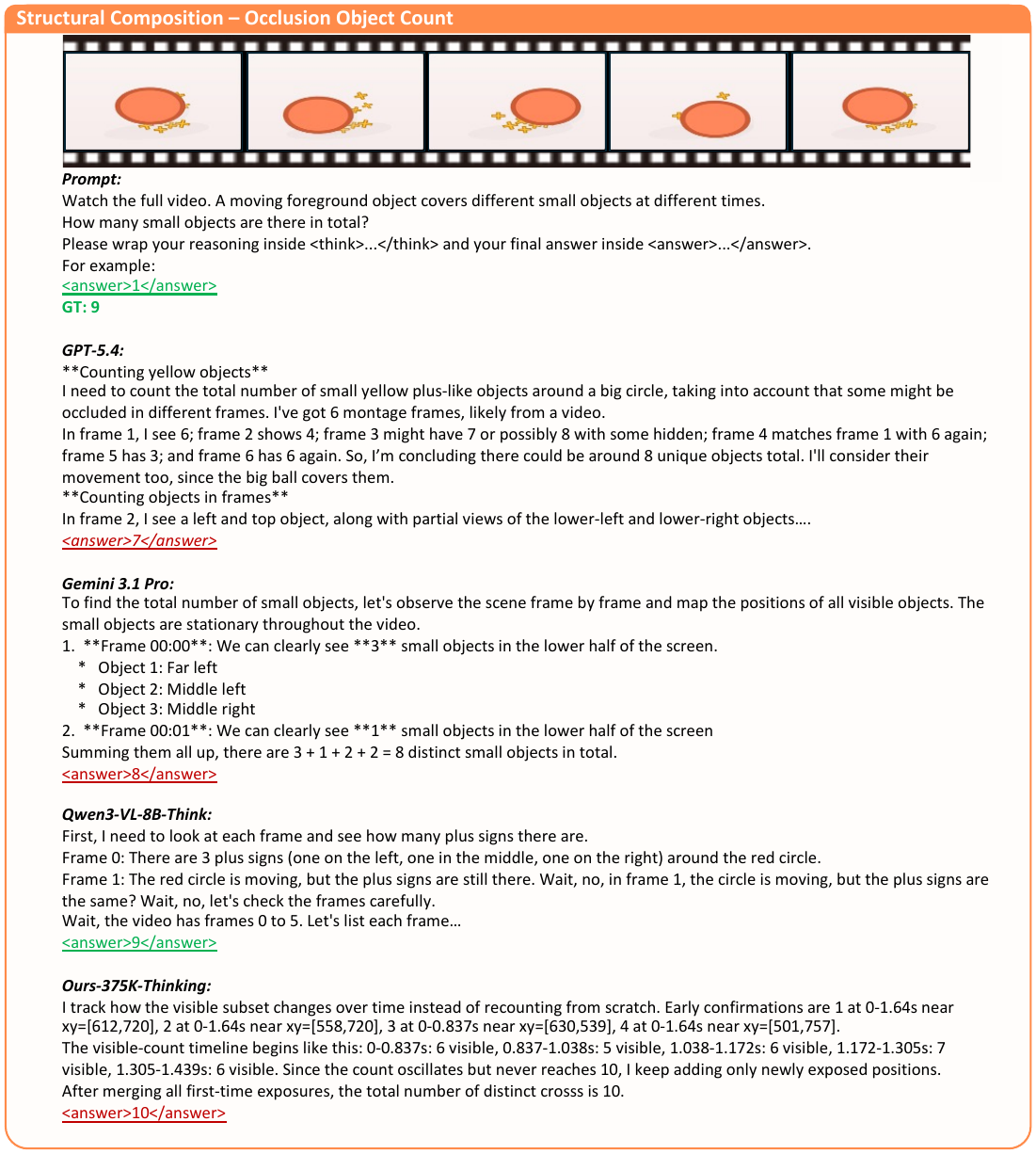}
  \caption{Additional visual examples from \textsc{Video-MME-Logical}.}
  \label{fig:appendix-show-data-03}
\end{figure*}

\begin{figure*}[h]
  \centering
  \includegraphics[width=0.96\textwidth,height=0.82\textheight,keepaspectratio]{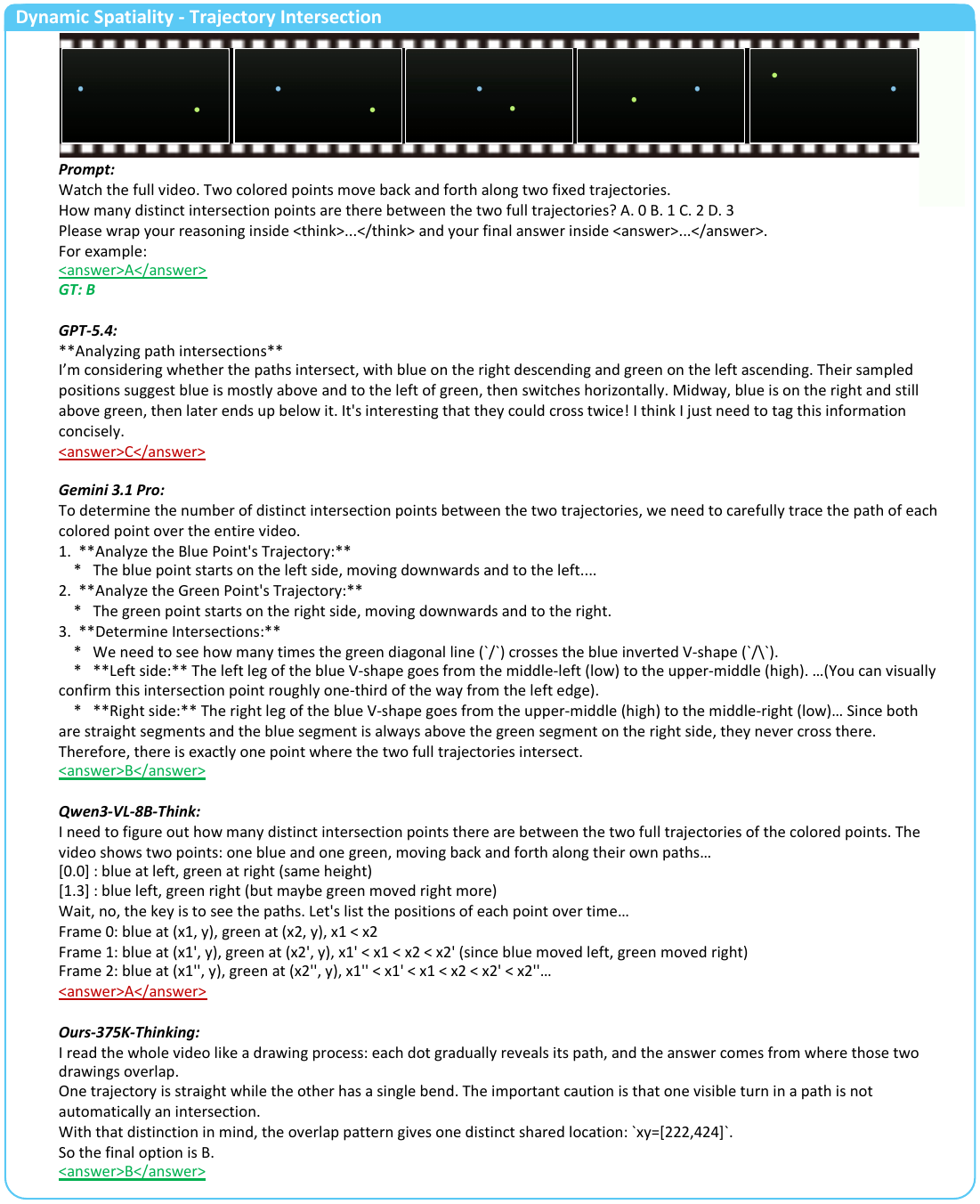}
  \caption{Additional visual examples from \textsc{Video-MME-Logical}.}
  \label{fig:appendix-show-data-04}
\end{figure*}

\begin{figure*}[h]
  \centering
  \includegraphics[width=0.96\textwidth,height=0.82\textheight,keepaspectratio]{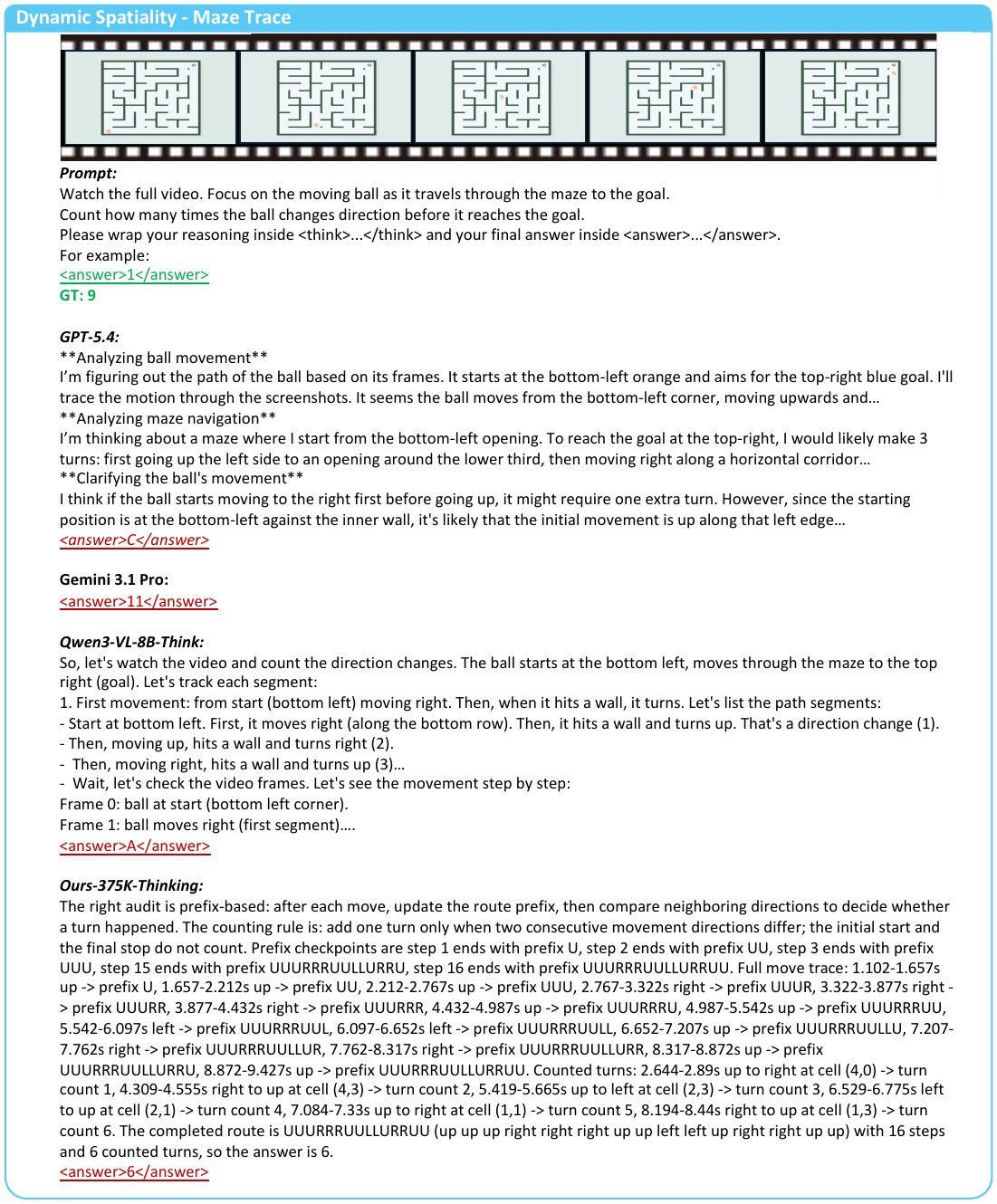}
  \caption{Additional visual examples from \textsc{Video-MME-Logical}.}
  \label{fig:appendix-show-data-05}
\end{figure*}

\begin{figure*}[h]
  \centering
  \includegraphics[width=0.96\textwidth,height=0.82\textheight,keepaspectratio]{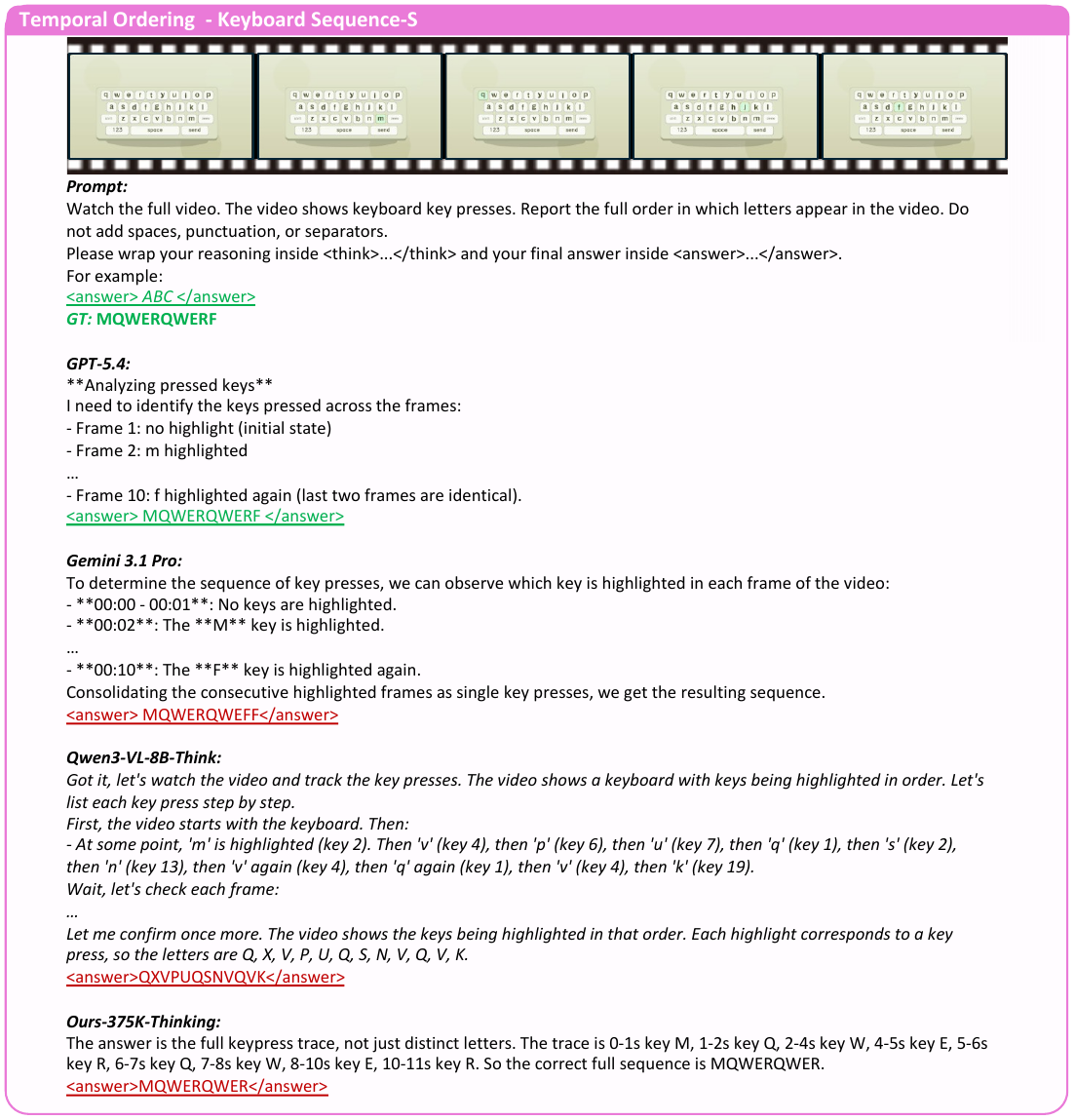}
  \caption{Additional visual examples from \textsc{Video-MME-Logical}.}
  \label{fig:appendix-show-data-06}
\end{figure*}

\begin{figure*}[h]
  \centering
  \includegraphics[width=0.96\textwidth,height=0.82\textheight,keepaspectratio]{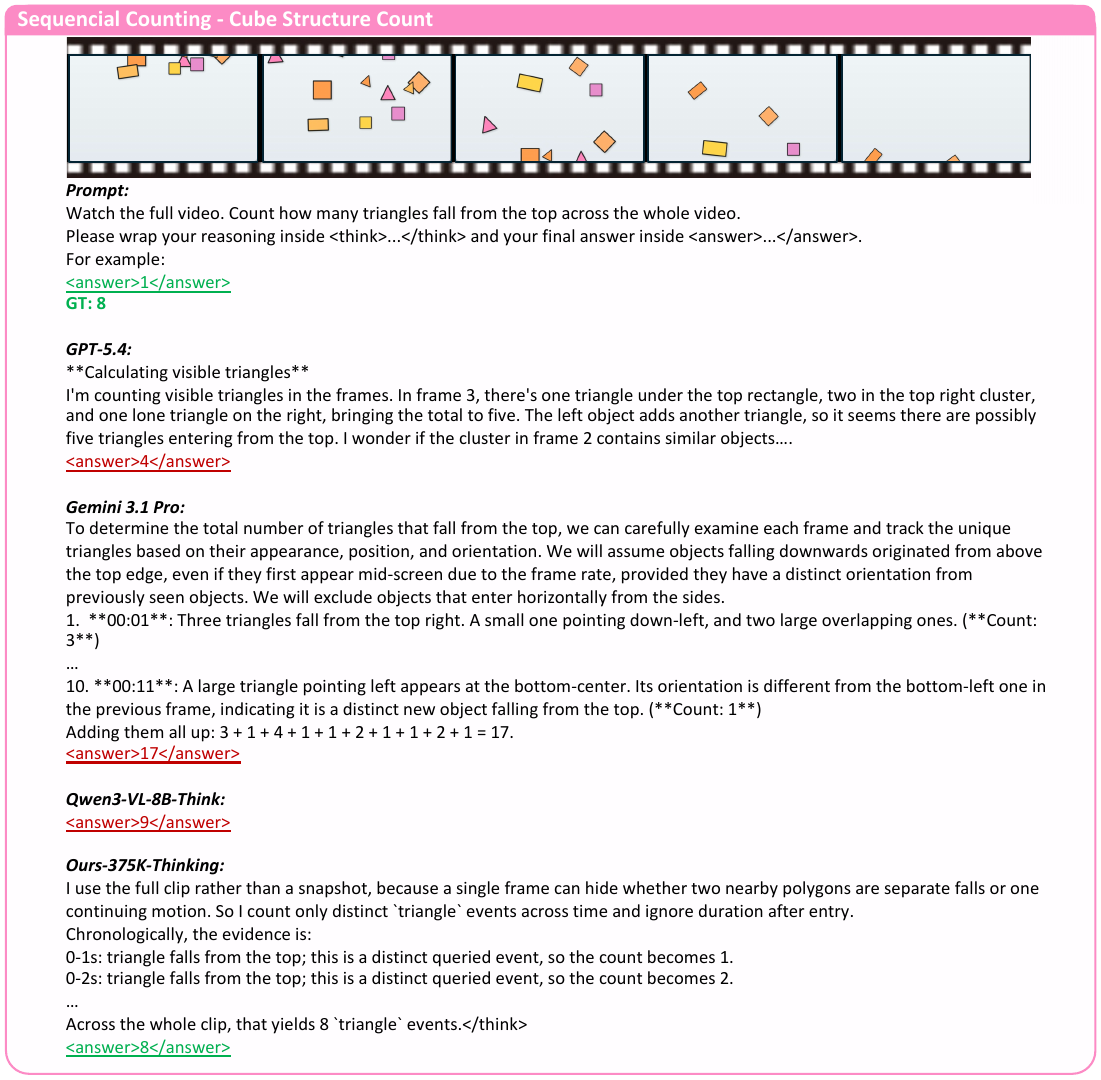}
  \caption{Additional visual examples from \textsc{Video-MME-Logical}.}
  \label{fig:appendix-show-data-07}
\end{figure*}

\end{document}